# Constraining Influence Diagram Structure by Generative Planning: An Application to the Optimization of Oil Spill Response

John M. Agosta
SRI International
333 Ravenswood Avenue
Menlo Park, CA 94025
johnmark@sri.com

## Abstract

This paper works through the optimization of a real world planning problem, with a combination of a generative planning tool and an influence diagram solver. The problem is taken from an existing application in the domain of oil spill emergency response. The planning agent manages constraints that order sets of feasible equipment employment actions. This is mapped at an intermediate level of abstraction onto an influence diagram. In addition, the planner can apply a surveillance operator that determines observability of the state—the unknown trajectory of the oil. The uncertain world state and the objective function properties are part of the influence diagram structure, but not represented in the planning agent domain. By exploiting this structure under the constraints generated by the planning agent, the influence diagram solution complexity simplifies considerably, and an optimum solution to the employment problem based on the objective function is found. Finding this optimum is equivalent to the simultaneous evaluation of a range of plans. This result is an example of bounded optimality, within the limitations of this hybrid generative planner and influence diagram architecture.

## 1 Introduction

Determining which actions to take and when has been addressed by distinct methods in AI. Among them are two techniques that choose actions in sequence to meet a specified goal. One is *generative planning*, coming out of the classical planning field; and the other, *influence diagrams*, out of the methods of Bayesian probability. Classical planning has its origins in symbolic reasoning methods created to address difficulties in conventional logic as facts change over time. Influence diagrams are Bayes' networks with decision and value nodes added, which were first introduced in the practice of decision analysis as an alternate representation for trees.

There have been attempts to find a sound extension of classical AI planning that encompasses uncertainty, or vice versa, to develop a decision theoretic planning method that structures the decision basis in addition to determining the best policies. Either extension must deal with the combinatorial complexity of generative planning on top of the complexity of the uncertainty calculation. [Blythe 1996] [Draper, Hanks and Weld 1994][Haddawy, Doan and Goodwin 1995][Lehner, Elaesser and Musman 1994][Lowrance and Wilkins 1990] The line of work presented here has a more modest goal; to demonstrate that the combination of planning and influence diagram methods can simplify the computational complexity of a real world problem by orders of magnitude. This solution exploits the abilities of the generative planner to manage constraints among sets of actions and the influence diagram to indicate irrelevant structure from the combinations of available actions, uncertainties and the objective function. The computational example presented in this paper suggests how the two methods in combination constitute an architecture for solution of complicated problems by working through a solution to an oil spill emergency response problem. On the one hand, the constraints generated by the planning agent determine the order and the range of actions incorporated in the structure that the influence diagram uses for optimization. On the other hand, the richness of the the influence diagram model can focus on which uncertainties and sets of actions are relevant, to further constrain the search of an optimum.

Beyond the practical value of this approach in solving real-world problems, this work has value in exploring the insight that as problems become more uncertain, the complexity of the planning problem can decrease rather than increase. The basis of this insight comes from the practice of modeling complex, uncertain problems in decision analysis, where the computational complexity of the models employed tends to be small. Decision analysis exploits properties of dominance and value of information, so that the model structure or computation generating informa-tion found to be irrelevant to the problem can be pruned.

My approach addresses similar concerns to methods of plan evaluation, in its ability to rank the quality of plans that have been generated by a planning agent. The term "plan evaluation" can also refer to measuring the success and computational performance of planning agents. A method that has been developed for plan evaluation in the first sense, called *action networks* [Goldszmidt and Darwiche 1994], has a strong similarity to the method applied in this paper. In contrast to action networks, influence diagrams represent decisions prior to deliberation and serve to optimize the choices at decisions. This also makes them capable of pre-posterior analysis, exploited in this paper, such as the computation of value of information, of which action networks are not capable. Of course, the lessons learned from action networks about representing time and persistence still apply.

The rest of this section reviews the model domain and architecture. Section 2 is a short summary of stochastic optimization. The step by step transformation and solution of the problem occupies Section 3, followed by conclusions in Section 4.

### 1.1 The U.S. Coast Guard Oil Spill Response Configuration System

We have developed a planning tool under contract with the U.S. Coast Guard (USCG) to assist them in the allocation and siting of the equipment available to clean up anticipated oil spills. As part of their required preparedness planning, the Spill Response Configuration System (SRCS) assists in determining the adequacy of the equipment to meet the threat of spills; for instance, either to determine to purchase more equipment, or, for that already staged at storage sites near the coast, to determine where best to site it. SRCS meets this objective by building plans against simulated oil spills to estimate the effectiveness of USCG's ability to clean them up. As intended in the original design, SRCS uses a generative planning tool, *System for Interactive Planning and Execution* (SIPE-2) [Wilkins 1988], to guide the user interactively through a set of progressively more detailed choices, first to choose cleanup strategies and then allocate and employ equipment against a chosen spill scenario. This paper describes an extension to SRCS to evaluate a range of plans generated by SIPE-2, and to choose the best.

### 1.2 The Agent Architecture

One approach to the limits that computation places on rational choice is to begin with a computational architecture—in this case a hybrid of two techniques—and optimize the solution method within the constraint of these architectural limitations on informational and computational resources. This extends the idea of "heuristic adequacy." Rather than beginning with the concept of a purely rational agent, and by approximation, whittling it down to make it tractable, this approach, termed *bounded optimality,* begins with the architectural assumption. Arguments for this approach to AI research, first made by Horvitz, are expanded in Russell's invited paper at the last IJCAI. [Russell 1995]

We may characterize a rational architecture by what computational and sensing abilities it possesses:

1. generative/evaluative: How much can the agent structure the problem?

2. static/dynamic: Can it reason about changes over time?

3. unobservable/observable: Is the state of the world accessible to the agent when it chooses?

4. deterministic/uncertain: Can it reason with partial or incomplete knowledge?

5. episodic/continuous control: Can it intervene only at selected times?

6. discrete actions/ continuous actions: Is the range of actions at any time finite or continuously variable?

In this application, generative planning is generative, dynamic, unobservable, deterministic, episodic and discrete. Influence diagrams are evaluative, dynamic, observable, uncertain, episodic and discrete. The combination includes three differences: generative versus evaluative, unobservable versus observable and deterministic versus uncertain. The comparison is perhaps unfair; the two methods purport to do different things. In generative planning, actions are distinguished by simply whether or not they achieve a goal, rather than the degree to which they achieve an objective. In comparison influence diagrams need to consider the degree of achievement of an objective to be able to optimize.

In looking to combine these methods I have taken a practical approach, asking which outputs of a mature planning tool, e.g. SIPE-2 can be used to structure an influence diagram that can then optimize the problem. As a stochastic optimization technique, influence diagrams solve a class of decision under uncertainty problems based on a quantitative objective, however the input to an influence diagram solver is a structure that has already determined the set of allowed actions and information observable in each episode, and the ordering of decision episodes. The challenge then, is to what extent can a generative planning tool create this linear ordering for the influence diagram, which I will call the *decision backbone.*

In the least commitment philosophy adopted by generative planning, the planning agent approaches the problem by a top down successive refinement method, terminating when the problem has been reduced to the lowest level procedures represented in its knowledge base. As the plan is refined, actions that pre-viously were parallel tend to become ordered. The process adds constraints to the set of actions, both by sequencing them in time, and by restricting the set of actions available at any one time. The process may be non-deterministic, in that if the problem is not fully constrained, the planner might arrive at any one of a number of satisfactory plans that meets the planning goals. In this work I don't examine exactly how the planner generates the plan; instead I examine which information is available to the planner, and how its constraints can be interpreted to structure the influence diagram for the problem. The answer in short, is first to use the planner output to sequence decisions in the influence diagram, and second to constrain sets of available actions at each decision point in the influence diagram. If the ordering of actions and the observations available at each set of actions are completely ordered, then the result is a single influence diagram. Actions to observe uncertain states are not typically encoded in a planner's set of operators although they are needed in the design of an influence diagram. To demonstrate the consequences of observations in a plan, the oil spill example has a surveillance operator that generates an observation action for the uncertain trajectory state.

In generative planning, refinement steps generate levels of the plan hierarchy that vary in their level of abstraction. As the planning agent applies operators to goals at one level, they are expanded into sub-plans at lower levels. It has been recognized that the ability to work at varying levels of abstraction offers a major simplification to the search problem. In contrast, the influence diagram works entirely at one level of abstraction. It is not my intent to make a contribution to the theory of abstraction per se, but, as with other features of generative planning, to show how abstraction can be exploited. The key to this is to recognize, at each decision point, which level of abstraction in the plan should be represented in the influence diagram.

This design decision depends on the level of the actions that have the most direct relevance to variables in the objective function, i.e.—that most directly correspond to the quantities of which the objective is a function. In this model these are the quantities of equipment employed in a sector. Other levels of the planning agent's goal-expansion operators are relevant to the objective only in how they constrain actions at the relevant level. They do not have a direct effect on the objective.

### 1.3 Concepts of value and goal achievement

The similarity between the notion of planning to meet a goal and of optimizing to improve an objective function hides the fact that they are represented differently, and serve different purposes in this architecture. A goal represents achievement of a binary condition; an objective represents a complete ordering of outcomes. (For an extended discussion on this see [Dean and Wellman 1991] section 5.6.) On the kind of problems with which classical planning started out, a binary condition as an objective is appropriate. These problems have all or nothing solutions, like jumping a chasm, or tossing a bean bag into a box. The value of the objective is high and constant if the bag falls in the box, or the chasm is breached, but low otherwise.

In this application, the planning goals and the objective function are used for different purposes in the architecture. The objective function of SRCS to be minimized is the expected quantity of oil that reaches the shore. Equivalently, this can be expressed as minimizing the fraction of oil that reaches the shore out of the total amount that would reach the shore if no actions were taken. The rich physical domain model expressed in the objective function that the optimization routine applies has no counterpart in the domain model of the planning agent. Even if there were a satisfactory way to dichotomize the objective outcomes into achievement and failure of a goal, this would not serve the purpose of using the planner to constrain the action sets of the influence diagram optimization. Instead, the planning goals represent the achievement of feasible actions to locate and transport equipment, at different levels of abstraction. The planning agent finds sets of actions that meet these goals, by considering, for example, what combinations of equipment are either possible or necessary at each location, what conditions were necessary before the equipment is brought to the location, or employed there, and what interactions, both helpful and harmful might exist between actions. The planner is underconstrained so that there exist many plans that meet these goals within the planning constraints. Optimization then takes place within this feasible set of actions that the planner has so determined.

## 2 Optimization under uncertainty

The general form of a stochastic optimization problem can be written:

$$V^* = \min_{P(o)} E_x[V(x, P(o))]$$

where $P(o)$ is a policy function of the subset, $o$, that can be observed of variables $x$, and $V()$ is a value function to be minimized. The world model is expressed as a joint distribution over all uncertain variables, $pr\{x|\xi\}$, where $\xi$ is the decision maker's state of knowledge before the first decision is made. In general $P(o)$ can be quite complicated. It can include a sequence of $i$ episodes, each containing a decision, where generally $\xi_i$ for each decision includes all observations and choices made in previous episodes. This set of decisions is represented by a completely ordered set of nodes, called the *decision backbone* of the influence diagram.

Decisions occur episodically at times called *decision*

*points.* At each decision point where the entire current state is observed, the problem can be factored, so that it can be solved in stages, and the policy at each stage is only a function of the current observed state. [Bertsekas 1987] This is shown in an influence diagram by a decision node, that, if removed—together with its uncertain direct predecessors—would split the decision backbone and disconnect the influence diagram network. The resulting solution is a recurrence equation over return functions $V_i$ known as Bellman's equation:

$$V_i(x) = \min_{p(x)} E_{x_{i+1}}[V_{i+1}(x_{i+1}, p(x))]$$

It can be applied sequentially, starting with the last stage, with $V_T$ set equal to the objective function, to solve the stochastic optimization problem, obtaining $V^* = V_0$. The complexity of this sequentially decomposable problem depends upon the size of the state space, $x_i$, of each stage. This can still be large. For instance, in the oil spill problem this state space must describe the two dimensional distribution of the oil slick at a point in time.

Bellman's equation applies to the solution of influence diagrams whose decisions meet the observability condition. To exploit this simplification, we will design structures where this condition occurs.

## 3  Solution of the best policy in the oilspill domain

### 3.1  The Nature of an Oil Spill Response

The problem addressed by SRCS is the emergency response to large oil spills caused by the foundering of commercial tankers or barges that transport of oil. Typically such a spill involves thousands of barrels of oil released over the course of several hours or days. The USCG and commercial carriers maintain on-call extensive inventories of oil spill response equipment in locations near ports where oil transport occurs. The expense to maintain this capability is small in comparison to the expected costs of damage and cleanup should a large spill reach shore.

By its nature, the oil spill cleanup problem is a race against time where the expanding extent of the threat due to the spreading oil slick is uncertain. The oil is dispersed by wind and tide in addition to advective spreading over the surface of the water. The purpose of the response is to prevent the oil from reaching sensitive areas on the shore by use of floating containment booms and removal methods such as oil skimmers, dispersant chemicals, or controlled burning.

The response problem is to determine which equipment should be dispatched when and where. Actions tend to concentrate 1) around the foundered ship, to prevent open water releases, 2) in open water, to remove the oil slick, and 3) at the shoreline, to protect sensitive areas. This categorization matches the highest level of abstraction of the planning operators. In this example we will restrict the actions considered to the use of booms and dispersants, without much loss of generality. For a specific incident, the location and weather often make it difficult to entirely control a spill discharge: Even with an optimal plan in this case the degree of containment and cleanup may be small. A detailed description of our formulation of this problem can be found in [Desimone and Agosta 1994].

### 3.2  Derivation of the Architecture: The Approach to Simplifying the Model

This section addresses the simplifications possible to the model based on the architecture of the planning and optimization tools. A somewhat extreme approach was taken here, more as a demonstration of what kind of transformations are possible by exploiting this hybrid architecture. Interestingly, within the bounds of the architecture, the optimality of the solution obtained is not significantly compromised. The only concession to simplifying assumptions made in the model was to reduce the number of sensitive areas threatened to 3, down from 6.

#### 3.2.1  Modeling of Trajectory Uncertainty

By borrowing from the modeling techniques used for oil trajectory modeling, [Spaulding 1988] significant reductions can be made in the size of the probability state space of the influence diagram. At the most general level, the dispersion of oil is a time dependent probability distribution of the amount of oil, $q$, over the two dimensional surface of the water: $pr\{q|u_1, u_2, t\}$. There are two common simplifications to this, by using either a Lagrangian frame or a Eulerian frame for discretization. In the Lagrangian formulation, which is standard practice for deterministic forecasting models, the quantity of spilled oil is discretized into "spillets" and each spillet's trajectory is forecast. Then spillets in an area are averaged to estimate oil quantities at each location. In comparison, in the Eulerian formulation, locations are discretized, and the quantities of oil are forecast on this grid. The disadvantage of the Eulerian formulation is that a diffusion process will propagate an infinitesimal quantity to each location, making it difficult to simplify the computation by localizing it. We show the compensating advantage to the Eulerian formulation when uncertainty is considered is that the geographic discretization can be coarse, so the state dimension remains small, and the uncertainty calculation can be piggy-backed on the oil diffusion calculation.

The Eulerian model is:

$$\mathbf{q}_{t+1} = M(\mathbf{q}_t),$$

where $\mathbf{q}_t$ is a vector of the oil quantities at locations in time $t$.

These transformations are made to the oil trajectory calculation:

Oil in the grid upon which the oil dispersion is calculated is aggregated up to the planning agent's level of abstraction. This divides the body of water into 6 sea-sectors and 6 shore-sectors. Thus the state space of the model, $x$, are the quantities of oil contained in sectors, expressed by the vector $\mathbf{s}$.

The trajectory model calculates the fraction of oil transported out of each sector to each adjoining sector instead of the total quantity. Thus transport can be represented as a Markov process, where the shore sectors are absorbing states for oil, with $M'_j$ as the row Markov matrix that describes the spreading process without the effect of uncertainty:

$$\mathbf{s}_{j+1} = M'_j \mathbf{s}_j$$

The effect of uncertainty in the wind and other physical forces on spreading is to make the position of the oil uncertain, in addition to spreading it wider. This is the major uncertainty faced in the oil spill response problem. The uncertainty in the oil location can be formulated as an additional spreading matrix, $P$, in the Markov process. To calculate the uncertain position of the oil, $M'_j$ is replaced by $M_j = PM'_j$

This is implemented by adding an uncertainty factor to the oil spreading rate when calculating $M_j$, and changing the interpretation of the calculated quantity $\mathbf{s}$ from the quantity of oil in a sector, to the $p$-th fractile $pr\{S < s\} = p$, where $S$ is a random variable and $p$ is a constant. This fractile will be denoated by $\mathbf{s}^p$.

Driven by the advantages of the Eulerian formulation this representation differs from that of the typical uncertain variables in an influence diagram in that the probability $p$ is discretized and quantity $\mathbf{s}$ is the variable value. Expected values can be calculated the same in either representation.

A further simplification is to use only one level of discretization, by setting $p = 0.5$. The result is that the same Markov calculation can be used for the spreading of the probability of oil as was used for oil. The oil's spreading factor is adjusted, and the interpretation changed to the output of the Markov process so that instead of $\mathbf{s}$ it generates $\mathbf{s}^{0.5}$ where the superscript refers to the value used for $p$.

### 3.3 The Objective Function

The computation of the quantity of oil that escapes collection by means of a Markov model is the entire objective function computation. The objective function calculation occurs over an array dimensioned by sectors and periods, where there is an account computed of the oil quantities for each one-hour-period and sector. Starting with the amount of oil entering a sector at the beginning of the period and the amount previously contained, this accounting determines 1) the amount of oil contained and removed by equipment in the sector during the period, 2) the amount transformed by natural and artificial processes, and 3) the amount free to escape to adjacent sectors—sea or land—in the next period. As described the movement of oil among sectors from period to period is determined by the Markov transition matrices. By the final period, in this case, 24 hours after the onset of the incident, substantially all the oil will have left the water's surface and come to rest, so its final disposition will be known.

Since the fraction of oil transported between sectors is assumed independent of the amount of oil in the sector, the set of transition matrixes $M_j$ can be pre-calculated for all periods, of the fraction of oil in each sector that transfers to adjoining sectors. This linearity property simplifies the computation by separating the calculation of the trajectory from equipment planning and the optimization computation, so that incrementally re-solving the trajectory model at each stage, depending up the equipment deployed is not necessary. [Agosta 1995]

The accounting and propagation calculation that determines the fate of the oil can be done for an allocation of equipment by:

$$\mathbf{s}_{j+1} = M_j[\mathbf{s}_j - \mathbf{e}_j(\mathbf{s}_j)]$$

where $\mathbf{s}_j$ is a vector of the amount of oil in each sector at time-step $j$; and $\mathbf{e}_j()$ is the vector-valued function to determine oil removed by equipment deployed at time $j$ in each sector. The time-steps indicate that evaluation is a discrete dynamic calculation, but the planner is ignorant of time-steps. Instead $\mathbf{e}_j()$ is determined by assigning equipment to periods based on the arrival times as determined by the planner. The objective, $V$, is a function only of $\mathbf{s}_T$, for final period, $T$.

The act of surveillance to observe the trajectory changes the state of knowledge of the trajectory at the subsequent decision point. At this point, there will be full knowledge of the location of the oil slick. This has the side effect of forcing a recomputation using $M'_j$, for periods 1 to $j$, the transition matrices for a spreading rate that does not include the increase in uncertainty of the oil location over time. (The prime signifies observation.) For purposes of implementation, $M'_j$ is also pre-calculated. The trajectory of quantities $\mathbf{s}^{0.5}$, of the median value when the oil position is uncertain is shown in Figure 1. The trajectory of actual oil position, $\mathbf{s}$ is shown in Figure 2.

### 3.4 Representation of the Spill and Response Goals in SIPE-2

SIPE-2 plans against the forecast of the trajectory. It has knowledge of:

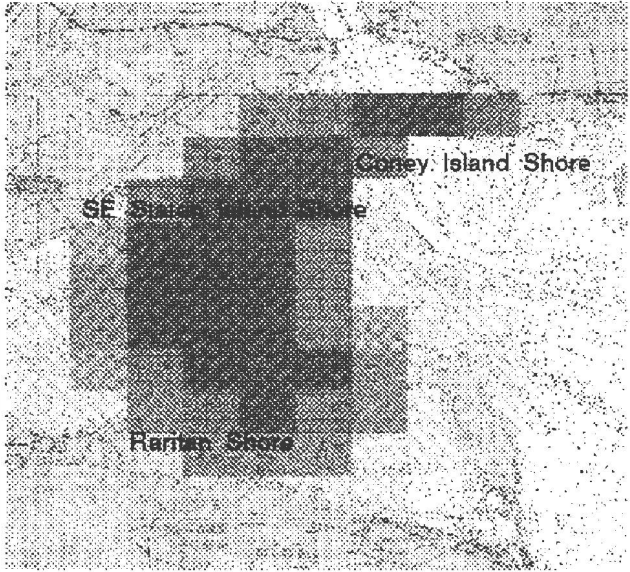

**Figure 1** The trajectory distribution before observation.

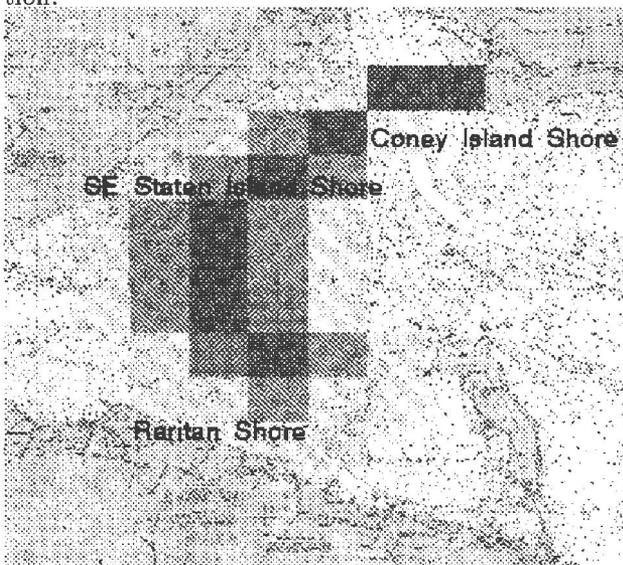

**Figures 2** The trajectory distribution after observation. Observation of the trajectory during surveillance has the effect of collapsing the known extent of the trajectory.

1. The spill rate and duration of the spill;
2. The location and probability of appearance of the oil, by geographic sector, s;
3. The extent (enclosing diameter) of the slick within a sector;
4. The thickness of the oil slick by sector, which affects the efficiency of oil removal.

SIPE-2 generates many possible plans of equipment employments to meet goals that derive from the description that it has of the trajectory. Each posted cleanup goal has a numeric value associated with it to suggest capacity of the equipment required. For example, to clean up a harbor, one may need enough length of floating boom to span the harbor entrance. When running the planning system interactively, the user takes these numeric levels as suggestions of adequate levels of response, around which the user is free to apply more equipment, or to neglect an employment goal entirely. In the combined architecture, the value of these quantities among the possible plans is determined by the influence diagram optimization step.

### 3.5 Model Simplifications Based on the Plan Output

Using SIPE-2 to construct the decision backbone from the generated plan applies the following constraints that lead to reformulations of the influence diagram and simplify its computation.

1) The arrival times determined by the planner completely order the employment actions, reducing a partial ordering of $2^n$ combinations of $n$ possible pieces equipment allocated to each location in each period, to a complete ordering.

2) It is always better for equipment to arrive in a sector sooner rather than later, so the dominant choice of boom for each sector is used to restrict the boom choice for each sector. If the minimum arrival times are such that no boom would arrive in time to have any material effect on one of the threatened sensitive areas, then no action to protect it will be taken.

3) The "friction" in moving equipment from one sector to another greatly reduces the opportunities for sequential action. The typical plan finds few reasons to move equipment once it has been deployed. Reuse takes time, and there are only two cases in the plan where it is feasible:

a) The "Observe or Disperse" decision. The choice is over which operation to perform first; to employ the one available aircraft for spreading chemical dispersants, or for surveillance on the course of the spill. The preparation and refueling times are so long that the decision reduces to either one of the other.

b) The "Chasing the spill" decision. As the spill progresses from the ship to the shore, should equipment

already employed somewhere else be moved in anticipation of its landfall?

Applying these constraints reduces the influence diagram model to two periods, where the later period has options for either disperse or observe the extent of the spill, and, subsequent to the observation, for relocating equipment. The complexity of the decision sequence has been reduced to the order of $2(2^n)$.

The codification of the USCG's best practices as represented in SIPE-2 plan operators add further constraints to the optimization:

1) Since booms leak, more boom is always better, but less than one times coverage of an area is useless (the oil will not be contained but just flow around), and three times coverage is the most practical. This further constrains the allocation of booms to sectors, since only boom lengths with integral converages are significant. This constrains the decision variable to a discrete value from the set $\{0, 1, 2, 3\}$, of the number of times of coverage of the area by boom. This sets the level of abstraction at which boom operations in the planner will be conveyed to the optimization computation. (It is possible that further planning to refine individual boom operations may be done in the planner after number of times coverage for each sector has been optimized.)

2) The earlier oil can be contained the better, since it spreads quickly.

These are the fundamental properties of the strategy that can be represented in the plan operators, without recourse to detailed properties of the physical and value models. They result in further reductions in the range of choices made available in the plan output to the influence diagram.

### 3.6 Solution results

Further simplifications occur in computational complexity due to the nature of the objective function.

1) An ounce of costs incurred in deploying equipment clearly outweighs the pounds of cleanup costs for damages, should the oil reach the shore. This eliminates the need to consider deployment cost consequences of the plan actions.

2) Of the boom combinations of the 4 booms available at the three locations to be protected, all but 3 are not dominated.

These three options for boom employment are:

1. *equal:* One layer of boom containment at the ship and at each sensitive area
2. *stabilize:* 3 layers of boom containment around the ship, at the sacrifice of the smaller sensitive area.
3. *(chase to) protect:* 3 layers of boom to protect the smaller sensitive area, leaving the ship's leakage

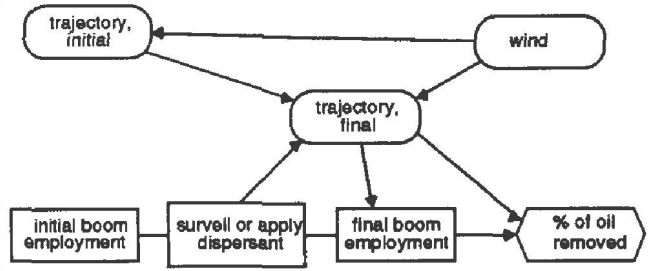

**Figure 3:** This influence diagram shows the two period structure of the simplified problem. The boom employment decisions contain three strategies: equal, stabilize, protect; with the addition of the chase strategy to the final boom decision. The observation is generated by choosing "surveillance" at the second decision. This changes the state of knowledge, indicated by the dependence of the trajectory probability distribution in the second period.

uncontained.

The last option must be split into two in the last period, since switching to "chase to" in the second period creates a delay in arrival time of the boom and thus makes it depend on the first period choice. At this point the structure of the influence diagram and the alternatives available at each decision are defined. The diagram is shown in Figure 3.

Three options in the first and four in the second period, together with the aircraft utilization choice leave $3 \times 2 \times 4 = 24$ plans to be evaluated by the influence diagram.

These can be further narrowed, by use of Bellman's recursion equation to solve the problem sequentially. The second period return function is

$$V_2(\mathbf{s}) = \min_{a, b_2} E_{s_{t'} \cdots s_T}[V(a, b_2, \mathbf{s})],$$

where $a$ is aircraft employment choice, and $b_2$ is the second boom employment choice. $V_2$ can be easily calculated:

| 2nd period action | aircraft | |
|---|---|---|
| | surveil | disperse |
| boom | | |
| equal | 0.29 | 0.19 |
| stablize | 0.60 | 0.17 |
| protect | 0.17 | **0.15** |
| chase | 0.46 | 0.22 |
| (none) | 1.00 | 0.47 |

**Table 1:** The fraction of oil left in sensitive areas after controls have been applied in the second period. The use-of-dispersants choice dominates the value gained from surveillance for all booming alternatives. The optimal policy is shown in boldface.

From the return function we have determined the best second period choices. There are two that are roughly tied, to either stablize or protect, while applying dispersants is always preferred. Strictly, the return func-tion should be calculated for all values of $s_{t'}$, and the values of $s_{t'}$ for each first period choice substituted into the return function. That would reduce the number of plan stages we need to evaluate to $8 + 3 = 11$. To verify this, Table 2 shows the computation of the final return (the objective function over all periods) for different combinations of first and second period choices. This shows that the choices determined for the second period remain valid when all periods are considered.

|  | 1st period action |  |  |
|---|---|---|---|
| 2nd period action | equal | stablize | chase |
| equal | 0.12 | 0.11 | * |
| stabilize | 0.05 | **0.03** | * |
| protect/chase | 0.10 | **0.03** | 0.10 |

**Table 2:** The fraction of oil left in sensitive areas for controls in both periods. Only the plans that choose use-of-dispersants are shown. The second period optimal choices remain optimal for the full calculation, as expected. The starred entries were not calculated, since they could be eliminated by best practice.

The two starred options, involve moving boom against the direction of the oil spreading (i.e. *protect* in the first period, and *switch* in the second) and would not be considered on basic physical grounds. This could also be included as a constraint during plan generation.

The final result is that *stabilize* in the first period, followed by either *stabilize* or *chase* in the second result in less than 3% of the oil reaching the shore that would have reached it had no controls been applied. The value of reducing uncertainty in the knowledge of the trajectory location by surveillance is minimal, and does not balance the reduction due to spreading of dispersants from the air. This says something about the "value of maneuver", to use the military metaphor, more than about the level of uncertainty in the oil trajectory. In essence, the spread of the oil is so wide that it will most likely reach all shores, and the limited ability to maneuver boom selectively to one or another turns out not to be important.

## 4   Conclusion

This paper has demonstrated an extreme example of how the daunting complexity of a real world domain can be made tractable by optimizing within the constraints of a combined generative planning and influence diagram architecture. The agent is rational only with respect to the computational limitations of the architecture, hence this result is an example of *bounded optimality*. Had a different architecture been applied, a different result might obtain.

The argument of the paper has been that the sophisticated design of the agent can take advantage of specifics of the problem domain to simplify the computation. The result is a more complicated architecture, but a simpler computation; in fact, in this example, one that approaches what can be done on a spreadsheet. The simplifications made within the constraints of the architecture are inspired by the characteristics of the domain. Finding them is gives a powerful tool: Perhaps not surprisingly, the richer description of the domain present in the combined architecture provides a better set of wedges that can be driven to split off large parts of the computation. The method applied is similar to "value of information" arguments; that, by knowing the available actions, their valuations and the states of knowledge under which they are taken, the constraints of the architecture do not significantly limit performance. In other terms, in cases with limited the flexibility, and limited range of options, the computational demands can be limited, even as the complexity and uncertainty faced may increase.

The missing link, and the task for further work is to show how the constraints that simplified the computational problem can be applied generally and automatically by the generative planner on the design of the influence diagram.

Acknowledgment: This work was partially supported under U.S.DOT grants DTRS-57-G-00084 and DTRS-57-94-G-00081.

## 5   References

Agosta, J. M. 1995. "Formulation and implementation of an equipment configuration problem with the SIPE-2 generative planner" In *Proceedings of the AAAI-95 spring symposium on integrated planning applications* Stanford, CA. March 1995.

Bertsekas, D. P. 1987. *Dynamic Programming: Deterministic and Stochastic Models*, Englewood Cliffs, NJ: Prentice Hall.

Blythe, J. 1996. "Event-Based Decompositions for Reasoning about External Change in Planners," To appear in *The Third International Conference on AI Planning Systems*, Edinburgh, May 1996.

Dean, T. and M. Wellman, 1991. *Planning and Control* San Mateo, CA: Morgan Kaufmann.

Desimone, R. V. and M. desJardins. 1993. "Representing and Reasoning About Uncertainty during Battle Planning" SRI Report, number ITAD-1549-FR-93-330.

Desimone, R.V. and J. M. Agosta. 1994. "Oil Spill Response Simulation: The Application of AI Planning Technology," *Proceedings of the 1994 Simulation MultiConference, Simulation for Emergency Management track,* La Jolla, CA, April 1994.

Draper, D., S. Hanks and D. Weld. 1994. "A Probabilistic Model of Action for Least-Commitment Planning with Information Gathering" In *Proceedings of the Tenth Conference on Uncertainty in Artificial Intelligence*, Seattle, August 1994.

Goldszmidt M. and A. Darwiche. 1994. "Action Networks: A Framework for Reasoning about Actions and Change under Uncertainty" In *ARPA/Rome Laboratory Knowledge-based Planning and Scheduling Initiative Workshop Proceedings*, Tucson, AZ, February, 1994.

Haddawy, P., A. Doan, R. Goodwin. 1995. "Efficient Decision-Theoretic Planning: Techniques and Empirical Analysis," In *Proceedings of the Eleventh Conference on Uncertainty In Artificial Intelligence* Montreal, Canada, August 1995.

Lehner, P., C. Elaesser and S. Musman. 1994. "Constructing Belief Networks to Evaluate Plans," In *AAAI Spring Symposium on Decision-Theoretic Planning*, Menlo Park, CA: AAAI Press.

Lowrance, J. D. and D. E. Wilkins, 1990. "Plan evaluation under uncertainty," in *Proceedings of the Workshop on Innovative Approaches to Planning, Scheduling and Control* (K. P. Sycara, ed.), pp. 439–449, Morgan Kaufmann Publishers Inc., San Francisco, CA, Nov. 1990.

Russell, S. 1995. "Rationality and Intelligence," *IJCAI - 95*, Montreal, August 1995.

Spaulding, M. 1988. "A State of the Art Review of Oil Spill Trajectory and Fate Modeling," *Oil and Chemical Pollution*, Vol. 4, pp. 39-55.

Wilkins, D. E. , K. L. Myers, J. D. Lowrance, and L. P. Wesley, 1995. "Planning and reacting in uncertain and dynamic environments," *Journal of Experimental and Theoretical AI*, vol. 7, no. 1, pp. 197–227,

Wilkins, D. E. 1988. *Practical Planning: Extending the Classical AI Planning Paradigm*, Morgan Kaufmann Publishers Inc., San Francisco, CA